\documentclass[letterpaper, 10 pt, conference]{class/ieeeconf}  

\IEEEoverridecommandlockouts                              
\usepackage{xcolor}
\usepackage{siunitx}  
\usepackage{graphicx}
 \graphicspath{./figures/robot_demonstration}

\usepackage{amsmath, bm}
\usepackage{amssymb}
\usepackage{latexsym}
\usepackage{url}
\usepackage{cite}
\usepackage{relsize}
\usepackage{multirow}
\usepackage{afterpage}
\usepackage{ifthen}
\usepackage{graphicx}
\usepackage{algpseudocode,algorithm,algorithmicx}
\usepackage{subfigure}
\usepackage{flushend}
\usepackage{epstopdf}
\usepackage{stackengine}
\usepackage{textcomp} 

\usepackage{gensymb}
\usepackage{booktabs}
\usepackage{makecell}
\usepackage{hhline}
\usepackage{courier}
\usepackage{lipsum}

\usepackage{xspace}
\usepackage{bbding}
\usepackage{tabularx}

\newcolumntype{C}[1]{>{\centering\arraybackslash}m{#1}}

\newcommand{\etal}{\textit{et al.}}

\makeatletter
\let\NAT@parse\undefined
\makeatother
\usepackage[bookmarks=false, linkcolor=blue, urlcolor=blue, citecolor=blue]{hyperref} 

\hypersetup{
    colorlinks=true,
    linkcolor=red,
    filecolor=magenta,      
    urlcolor=blue,
    pdfstartview={FitH},
    citecolor =blue
    }

\title{\LARGE \bf Language-Conditioned Affordance-Pose Detection in 3D Point Clouds}

\author{Toan Nguyen$^{1}$, Minh Nhat Vu$^{2}$, Baoru Huang$^3$, Tuan Van Vo$^{1}$, Vy Truong$^{1}$, Ngan Le$^4$ \\Thieu Vo$^5$, Bac Le$^6$, Anh Nguyen$^7$
\thanks{$^1$ FPT Software AI Center, Vietnam {\tt toannt28@fpt.com}}
\thanks{$^2$ Automation \& Control Institute, TU Wien, Vienna, Austria 
}
\thanks{$^3$ Imperial College London, UK
}
\thanks{$^4$ Department of Computer Science \& Computer Engineering, University of Arkansas, USA 
}
\thanks{$^5$ Faculty of Mathematics and Statistics, Ton Duc Thang University, Ho Chi Minh City, Vietnam} 
\thanks{$^6$ Department of Computer Science, University of Science, Vietnam}
\thanks{$^7$ Department of Computer Science, University of Liverpool, UK 
}}

\begin{document}

\newtheorem{problem}{Problem}
\newtheorem{lemma}{Lemma}
\newtheorem{theorem}[lemma]{Theorem}
\newtheorem{claim}{Claim}
\newtheorem{corollary}[lemma]{Corollary}
\newtheorem{definition}[lemma]{Definition}
\newtheorem{proposition}[lemma]{Proposition}
\newtheorem{remark}[lemma]{Remark}
\newenvironment{LabeledProof}[1]{\noindent{\it Proof of #1: }}{\qed}

\def\beq#1\eeq{\begin{equation}#1\end{equation}}
\def\bea#1\eea{\begin{align}#1\end{align}}
\def\beg#1\eeg{\begin{gather}#1\end{gather}}
\def\beqs#1\eeqs{\begin{equation*}#1\end{equation*}}
\def\beas#1\eeas{\begin{align*}#1\end{align*}}
\def\begs#1\eegs{\begin{gather*}#1\end{gather*}}

\newcommand{\poly}{\mathrm{poly}}
\newcommand{\eps}{\epsilon}
\newcommand{\e}{\epsilon}
\newcommand{\polylog}{\mathrm{polylog}}
\newcommand{\rob}[1]{\left( #1 \right)} 
\newcommand{\sqb}[1]{\left[ #1 \right]} 
\newcommand{\cub}[1]{\left\{ #1 \right\} } 
\newcommand{\rb}[1]{\left( #1 \right)} 
\newcommand{\abs}[1]{\left| #1 \right|} 
\newcommand{\zo}{\{0, 1\}}
\newcommand{\zonzo}{\zo^n \to \zo}
\newcommand{\zokzo}{\zo^k \to \zo}
\newcommand{\zot}{\{0,1,2\}}
\newcommand{\en}[1]{\marginpar{\textbf{#1}}}
\newcommand{\efn}[1]{\footnote{\textbf{#1}}}
\newcommand{\vecbm}[1]{\boldmath{#1}} 
\newcommand{\uvec}[1]{\hat{\vec{#1}}}
\newcommand{\thv}{\vecbm{\theta}}
\newcommand{\junk}[1]{}
\newcommand{\var}{\mathop{\mathrm{var}}}
\newcommand{\rank}{\mathop{\mathrm{rank}}}
\newcommand{\diag}{\mathop{\mathrm{diag}}}
\newcommand{\tr}{\mathop{\mathrm{tr}}}
\newcommand{\acos}{\mathop{\mathrm{acos}}}
\newcommand{\atantwo}{\mathop{\mathrm{atan2}}}
\newcommand{\SVD}{\mathop{\mathrm{SVD}}}
\newcommand{\quadf}{\mathop{\mathrm{q}}}
\newcommand{\linterp}{\mathop{\mathrm{l}}}
\newcommand{\sgn}{\mathop{\mathrm{sign}}}
\newcommand{\sym}{\mathop{\mathrm{sym}}}
\newcommand{\avg}{\mathop{\mathrm{avg}}}
\newcommand{\mean}{\mathop{\mathrm{mean}}}
\newcommand{\erf}{\mathop{\mathrm{erf}}}
\newcommand{\grad}{\nabla}
\newcommand{\R}{\mathbb{R}}
\newcommand{\defeq}{\triangleq}
\newcommand{\dims}[2]{[#1\!\times\!#2]}
\newcommand{\sdims}[2]{\mathsmaller{#1\!\times\!#2}}
\newcommand{\udims}[3]{#1}
\newcommand{\udimst}[4]{#1}
\newcommand{\com}[1]{\rhd\text{\emph{#1}}}
\newcommand{\ind}{\hspace{1em}}
\newcommand{\argmin}[1]{\underset{#1}{\operatorname{argmin}}}
\newcommand{\floor}[1]{\left\lfloor{#1}\right\rfloor}
\newcommand{\step}[1]{\vspace{0.5em}\noindent{#1}}
\newcommand{\quat}[1]{\ensuremath{\mathring{\mathbf{#1}}}}
\newcommand{\norm}[1]{\left\lVert#1\right\rVert}
\newcommand{\ignore}[1]{}
\newcommand{\specialcell}[2][c]{\begin{tabular}[#1]{@{}c@{}}#2\end{tabular}}
\newcommand*\Let[2]{\State #1 $\gets$ #2}
\newcommand{\algorithmicbreak}{\textbf{break}}
\newcommand{\Break}{\State \algorithmicbreak}
\newcommand{\ra}[1]{\renewcommand{\arraystretch}{#1}}

\renewcommand{\vec}[1]{\mathbf{#1}} 

\algdef{S}[FOR]{ForEach}[1]{\algorithmicforeach\ #1\ \algorithmicdo}
\algnewcommand\algorithmicforeach{\textbf{for each}}
\algrenewcommand\algorithmicrequire{\textbf{Require:}}
\algrenewcommand\algorithmicensure{\textbf{Ensure:}}
\algnewcommand\algorithmicinput{\textbf{Input:}}
\algnewcommand\INPUT{\item[\algorithmicinput]}
\algnewcommand\algorithmicoutput{\textbf{Output:}}
\algnewcommand\OUTPUT{\item[\algorithmicoutput]}

\maketitle
\thispagestyle{empty}
\pagestyle{empty}

\begin{abstract}

Affordance detection and pose estimation are of great importance in many robotic applications. Their combination helps the robot gain an enhanced manipulation capability, in which the generated pose can facilitate the corresponding affordance task. Previous methods for affodance-pose joint learning are limited to a predefined set of affordances, thus limiting the adaptability of robots in real-world environments. In this paper, we propose a new method for language-conditioned affordance-pose joint learning in 3D point clouds. Given a 3D point cloud object, our method detects the affordance region and generates appropriate 6-DoF poses for any unconstrained affordance label. Our method consists of an open-vocabulary affordance detection branch and a language-guided diffusion model that generates 6-DoF poses based on the affordance text. We also introduce a new high-quality dataset for the task of language-driven affordance-pose joint learning. Intensive experimental results demonstrate that our proposed method works effectively on a wide range of open-vocabulary affordances and outperforms other baselines by a large margin. In addition, we illustrate the usefulness of our method in real-world robotic applications. Our code and dataset are publicly available at~\href{https://3dapnet.github.io}{https://3DAPNet.github.io}.

\end{abstract}


\section{INTRODUCTION} \label{Sec:Intro}


In robotic research, affordance detection and pose estimation are among the most important and well-concerned problems~\cite{hassanin2021visual,duan2021robotics}. Understanding object affordance helps robots decide the inherent possibilities and potential actions within an environment, while pose estimation is considered a prerequisite for robots to interact with and manipulate their surrounding objects effectively. Combining affordance detection and pose estimation holds the potential to help robots gain a more comprehensive understanding of their environment's possibilities and, at the same time, achieve enhanced manipulation abilities~\cite{newbury2023deep}. However, prior research has predominantly focused on solving these problems independently, while less works tackled both tasks simultaneously~\cite{liu2020cage,chen2022learning,guo2023handal}. This is because the concept of affordance can be arbitrary, and without extra information (e.g., text input), it is challenging to detect the associated pose.
\begin{figure}
\centering
\includegraphics[width=0.85\linewidth]{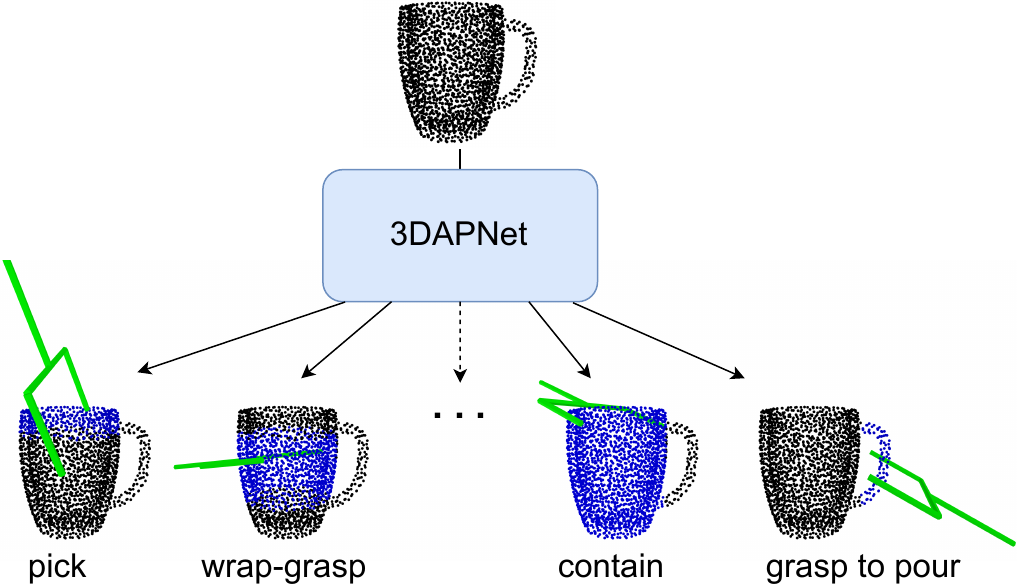}
\vspace{0.25pc}
\caption{Our framework allows the simultaneous detection of affordance region and corresponding supporting poses given the input point cloud object and an arbitrary affordance text.}
\label{fig:intro}
\end{figure}

Recently, with the availability of depth cameras, several works have addressed the task of affordance detection in 3D point clouds~\cite{kim2015interactive,deng20213d,mo2022o2o,chen2022learning,nguyen2023open}. Most of them treated the problem as a supervised task of labeling predefined affordance labels for each point in the point cloud~\cite{deng20213d,chen2022learning}. Lately, the authors in~\cite{nguyen2023open} explored the open-vocabulary affordance detection task, a new research direction liberating the constraint of a predefined affordance label set with the utilization of language models~\cite{radford2021learning,devlin2018bert}. The work in~\cite{nguyen2023open} increased the flexibility of the affordance learning process, getting closer to universal affordance detection, however, it does not provide 6-DoF poses that supports the corresponding affordance. As a result, the task remains a visionary problem and currently hinders its practical application on real robots. Other works exhibit a combination of affordance detection and pose estimation~\cite{kokic2017affordance,zhao2022dualafford,chen2022learning,he2023pick2place}, yet their methods are still limited to a predefined set of affordance tasks.


In this research, we take a step further by integrating the tasks of \textit{open-vocabulary} affordance detection and \textit{language-driven pose estimation}. Given a 3D point cloud, our goal is to simultaneously detect the unconstrained affordance and generate poses based on the input text query. To realize that objective, we first establish a new dataset for the task of 3D Affordance-Pose joint learning, namely 3DAP dataset. Our dataset is composed of several triplets of a 3D point cloud, an affordance label in the form of the natural text, and a set of multiple 6-DoF poses associated with the affordance.
We then present a joint learning framework consisting of a language-driven affordance detection branch and a pose estimation branch which is a guided diffusion model that generates 6-DoF poses conditioned on the given point cloud object and the affordance text. Our choice of the diffusion model is motivated by its recent remarkable results in generating diverse data modalities from multiple conditions~\cite{luo2021diffusion,li2022diffusion,blattmann2023align}, yet its application to pose estimation remains limitedly explored~\cite{urain2023se}. 
Our method is an end-to-end pipeline where via a text prompt, the robot can perform a manipulation task using the affordance and the detected pose. Figure~\ref{fig:intro} shows the main concept of our work.

Our contributions are summarized as follows:
\begin{itemize}
    \item We introduce 3DAP, a new dataset of 3D point cloud objects with affordance language labels and affordance-specific 6-DoF poses.
    \item We propose 3DAPNet, a new method that effectively tackles the task of affordance-pose joint learning.
    \item We validate our method through intensive experiments and demonstrate the usefulness of 3DAPNet in several real-world robotic manipulation tasks.
\end{itemize}
\section{Related Work} \label{Sec:rw}

\textbf{Affordance Detection.}
Many works tackled the task of affordance detection in RGB images~\cite{nguyen2017object,do2018affordancenet,luo2023leverage,chen2023affordance,mur2023bayesian} and 3D point clouds~\cite{deng20213d,tuanopenkd,mo2022o2o,chen2022learning,nguyen2023open}. 
In particular, Luo~\etal~\cite{luo2023leverage} leveraged affinity from human-object interaction to detect affordances of non-interactive objects in 2D images. Authors in~\cite{kim2015interactive} detected affordance maps on 3D point cloud scenes through interactive manipulation. Also working on point clouds, authors in~\cite{mo2022o2o} proposed a framework that detects affordance maps from object-object interaction. Most of these works focus on detecting a set of predefined affordances rather than open-vocabulary setting. Lately, Nguyen~\etal~\cite{nguyen2023open} introduced a framework that allows the detection of arbitrary affordance given in form of a text description. While achieving promising results, the common shortcoming of previous methods is that they solely detect the affordance regions while usually neglecting the corresponding poses that support the detected affordances. This limitation poses a challenge for the robot to effectively execute necessary affordance tasks in real-world manipulation settings. 

\textbf{3D Pose Generation.}
Given a single object or multiple objects in a cluttered environment, the goal of 3D pose generation algorithms is to find a pose configuration that can support manipulation tasks~\cite{duan2021robotics,newbury2023deep}.
Initial works addressed the problem by employing analytical approaches~\cite{nguyen1988constructing,bicchi2000robotic}, which are practically limited since they assume complete knowledge of object properties like shape, geometry, and material. The rapid development of grasping simulators~\cite{miller2004graspit,koenig2004design,todorov2012mujoco} in the following years led to the rise of data-driven approaches. Early data-driven methods primarily used whether hand-crafted features~\cite{kyota2005detection,michel2006approach,el2007learning} or traditional machine learning algorithms~\cite{saxena2008learning,saxena2008robotic,jiang2011efficient}. Recently witnessed the groundbreaking performances of deep learning methods for pose estimation~\cite{nguyen2016preparatory,mousavian20196,lou2021collision,chen2022learning,sen2023scarp, schiavi2023learning}. In particular, Lou~\etal~\cite{lou2021collision} presented a method that can generate collision-free poses in challenging environments, while authors in~\cite{mousavian20196} synthesized poses using a variational autoencoder network. With the recent remarkable results in various generation tasks, diffusion models have also been applied for the task of pose generation~\cite{huang2023diffusion,urain2023se}. Different from these approaches which are affordance-agnostic, our proposed diffusion model tackles the task of affordance-specific pose generation. Some other earlier works leveraged the affordance learning for the problem of task-specific grasping~\cite{kokic2017affordance,zhao2022dualafford,chen2022learning}. Nonetheless, their methods are restricted to a predetermined set of affordance tasks. In comparison, our method focuses on the open-vocabulary setting.


\textbf{Language-Conditioned Robotic Manipulation.}
With the stunning advancements of large language models~\cite{devlin2018bert,brown2020language,chowdhery2022palm,touvron2023llama,driess2023palm,vuong2023graspanything}, several recent works~\cite{venkatesh2021translating,silva2021lancon,ahn2022can,garg2022lisa,mees2023grounding,ren2023leveraging,tang2023task} have utilized the rich semantics of language for the tasks of robotic manipulation. For instance, Ahn~\etal~\cite{ahn2022can} proposed a method that constrains the language model to recommend actions that are both plausible to the robot and contextually appropriate. Silva~\etal~\cite{silva2021lancon} proposed to use language to support generalization in multi-task manipulation. The authors in~\cite{garg2022lisa} presented a framework that can learn meaningful skill abstractions from language-based expert demonstrations. More recently, Ren~\etal~\cite{ren2023leveraging} introduced a language-conditioned and meta-learning approach that learns efficient policies adaptable to novel tools from text descriptions. Different from these works, our method addresses the task of language-conditioned affordance-pose joint learning, where the affordance language simultaneously grounds the affordance region and 6-DoF pose configurations. 
\section{The 3D Affordance-Pose Dataset} \label{Sec:dataset}

\begin{figure}
\centering
\includegraphics[width=0.95\linewidth]{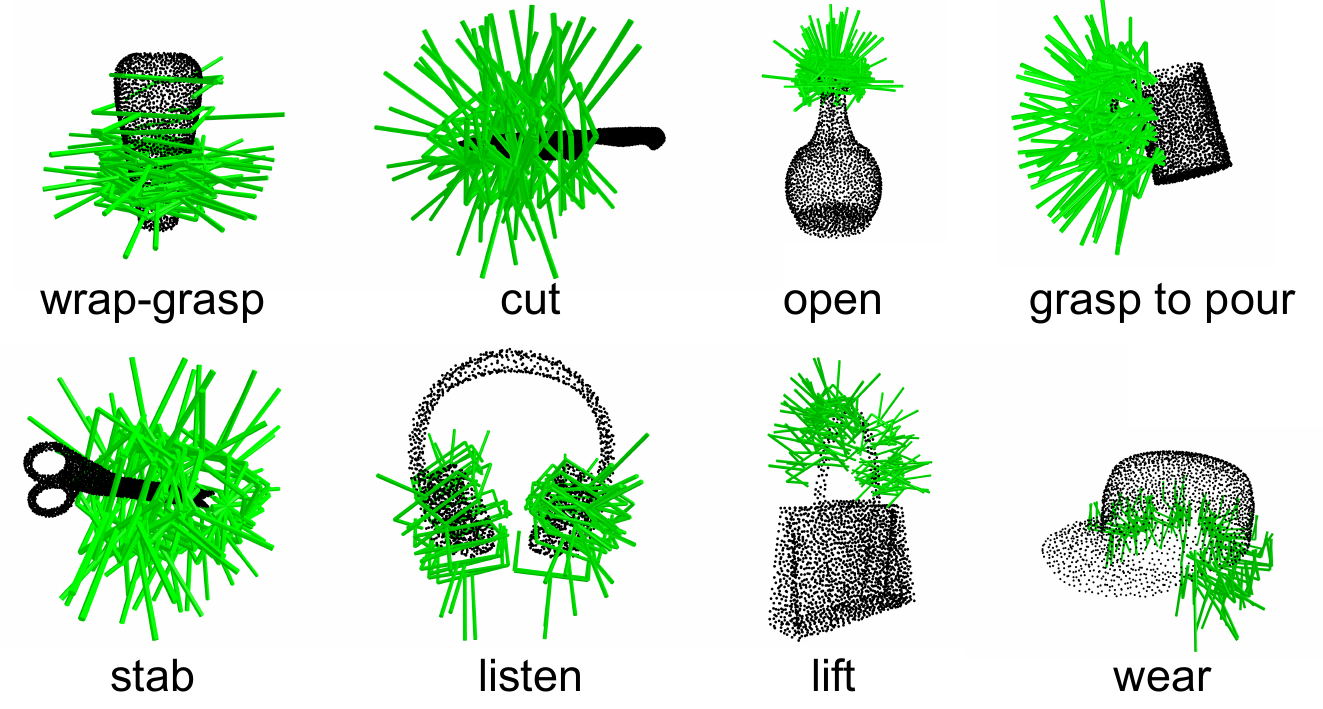}
\vspace{1ex}
\caption{Affordance-specific pose examples.}
\label{fig:poses}
\end{figure}

We present the 3D Affordance-Pose dataset (3DAP) as a dataset for affordance-pose joint learning. To construct this dataset, we apply a semi-automatic pipeline in which we first collect affordance-annotated 3D point clouds from 3D AffordanceNet~\cite{deng20213d}, a widely-used and currently the largest dataset for affordance detection in 3D point clouds. Next, we leverage 6-DoF GraspNet~\cite{mousavian20196} method to generate a large number of 6-DoF pose candidates. Afterwards, we manually select the affordance-specific poses for each affordance that the object affords. 

\begin{figure*}[t]
	\centering
	\includegraphics[width=0.85\linewidth]{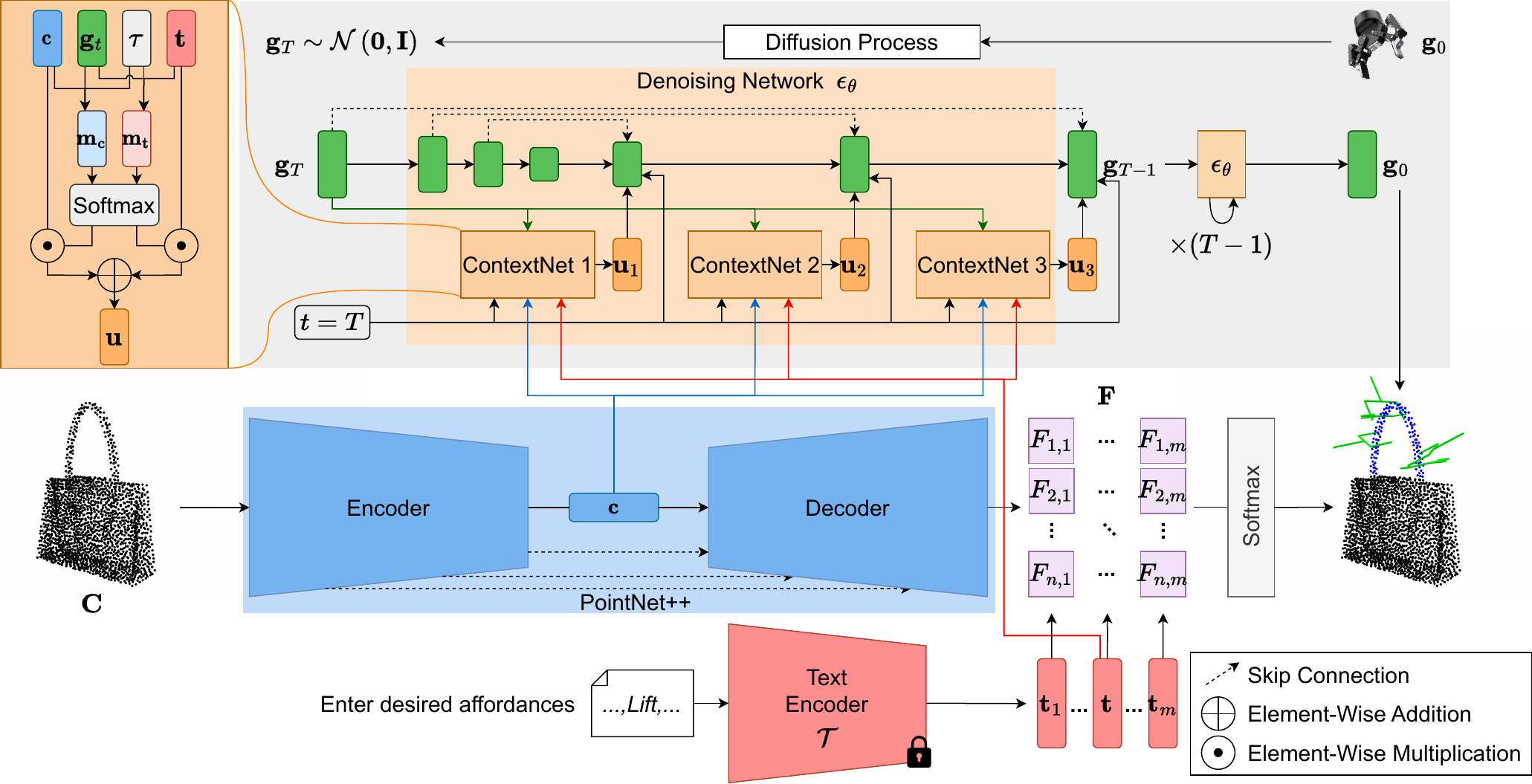}
 \vspace{1.5ex}
	\caption{The overview of our 3DAPNet. Our network includes two branches: one for affordance detection and one for pose generation. The unrestricted affordance label represented in natural language form enables the open-vocabulary setting. In inference, the predicted affordance map and the generated pose are combined to further support the appropriate task. 
 }
	\label{fig:architecture}
\end{figure*}

\textbf{Point Cloud Collection.}
We collect affordance-annotated point clouds from the recent 3D AffordanceNet dataset~\cite{deng20213d}. Each point cloud represents a single object and is an unorder set of $2,048$ points. Each point is represented by its Euclidean coordinate. The point coordinate of every point cloud are normalized to be in $[0, 1]$. In order to well represent the real-world objects, we scale the point clouds by different scale factors so that the longest side of an axis-aligned bounding box for each object is from \SI{5}{\cm} to \SI{30}{\cm}. The collected objects are of well-used categories in the daily manipulation tasks, such as \texttt{knife}, \texttt{bottle}, or \texttt{mug}, etc.
We express affordance labels as natural language descriptors. This facilitates open-vocabulary affordance detection, so that methods trained on our 3DAP dataset can potentially generalize to unseen affordances. 

\textbf{Poses Collection.}
We utilize 6-DoF GraspNet~\cite{mousavian20196} to automatically generate a large number of pose candidates for each collected point clouds. In particular, for each object, we pick $1,000$ successful parallel-jaw poses with highest evaluating scores. Following the Robotiq 2F-85 setting~\cite{2f85}, the collected poses have the maximum grip aperture of \SI{85}{\milli\meter}. 
From the generated poses, we manually select the affordance-specific poses for each object. Given an object and an affordance, we select among $1,000$ candidates ones that best support the affordance task. For example, with a \texttt{bottle} and affordance \texttt{open}, the poses whose contact points lie on the lid are curated. In total, our dataset contains 28K gripper poses for a wide variety of affordance tasks. Examples of affordance-specific poses in our dataset are presented in Figure~\ref{fig:poses}.

\section{Affordance-Pose Joint Learning} \label{Sec:method}

\subsection{Problem Formulation}
We present 3DAPNet, a new method for affordance-pose joint learning. Given the captured 3D point cloud of an object and a set of affordance labels conveyed through natural language texts, our objective is to jointly produce both the relevant affordance regions and the appropriate pose configurations that facilitate the affordances. Particularly, 3DAPNet takes as input a point cloud denoted by $\mathbf{C} = \left \{\mathbf{p}_1, \mathbf{p}_2,...,\mathbf{p}_n \right \}$, containing $n$ points in 3D Euclidean space, alongside $m$ arbitrary affordance labels articulated in natural language. The desired output from our framework encompasses an affordance map $\mathbf{A}=\left \{a_1, a_2,...,a_n \right \}$ that assigns an affordance label to each point
, and $m$ sets of 6-DoF poses that facilitate corresponding affordances. We consider a 6-DoF pose as configured by $\left [\mathbf{g}_{\text{qu}}, \mathbf{g}_{\text{tr}} \right ]$, in which $\bf{g}_{qu}$ is a unit-norm quaternion representing the rotation and $\bf{g}_{tr}$ is a translation vector. The overview of our network is illustrated in Figure~\ref{fig:architecture}.



\subsection{Open-Vocabulary Affordance Detection}
We follow the recent work~\cite{nguyen2023open} to detect affordances with open-vocabulary setting. The input point cloud $\mathbf{C}$ is plugged into a PointNet++ model~\cite{qi2017pointnet++} to extract $n$ point-wise feature vectors $\mathbf{P}_1, \mathbf{P}_2,...,\mathbf{P}_n$. Next, the $m$ affordance language labels are fed into a text encoder $\mathcal{T}$ to extract $m$ text embeddings $\mathbf{t}_1,\mathbf{t}_2,...,\mathbf{t}_m$. Similar to other works~\cite{nguyen2023open,xu2023side,luo2023segclip}, the choices for the text encoder are versatile.

To enable open-vocabulary affordance detection, we compute the semantic relations between the point cloud affordance and its potential labels by correlating the text embeddings and point features using cosine similarity function. Concretely, the $F_{i,j}$  element at the $i$-th row and $j$-th column of the correlation matrix $\mathbf{F} \in \mathbb{R}^{n\times m}$, which is the correlation score of the point feature $\mathbf{P}_{i}$ and the affordance text embedding $\mathbf{t}_{j}$, is computed as:
\begin{equation}
{F}_{i,j} = \frac{{\mathbf{P}^\top_{i}} \mathbf{t}_j}{\left \| \mathbf{P}_i \right \|\left \| \mathbf{t}_j \right \|}\:\:.
\end{equation}

During the training, we optimize the PointNet++ to provide point embeddings that are close to the corresponding label text embeddings. The point-wise softmax output of every point $i$ is then computed as:
\begin{equation}
    {S}_{i,j}= 
    \dfrac{\exp\left ({F}_{i,j}/\eta\right )}
    {\sum_{k=1}^{m}\exp({{F}_{i,k}}/\eta)}\:\:,
\end{equation}
where $\eta$ is a learned parameter.
The loss function for affordance detection is computed as the negative log-likelihood of the softmax output over the entire point cloud:
\begin{equation}
    \mathcal{L}_{\text{aff}} = -\sum_{i=1}^n\log {S}_{i,{a_i}}.
\end{equation}

\subsection{Language-Conditioned Pose Generation}
Our key contribution is a new guided diffusion model to address the task of affordance-specific pose generation. Our diffusion model is designed to produce poses that not only based on the point cloud, but also facilitate the affordance task by conditioning on the input text. 

\textbf{Forward Process.}
Given a pose from the dataset $\mathbf{g}_0 \sim q(\mathbf{g})$, in the forward process, we gradually add to the pose small amounts of Gaussian noise in $T$ steps, creating a sequence of noisy poses $\textbf{g}_1, \: \textbf{g}_2,\ldots, \: \textbf{g}_T$. When $T\rightarrow \infty$, $\mathbf{g}_T$ is equivalent to $\mathcal{N}\left (\mathbf{0}, \mathbf{I} \right )$~\cite{ho2020denoising}. The noise step sizes are specified by a predefined variance schedule $\left \{ \beta_t \in (0, 1)\right \}^T_{t=1}$. From that, the forward process is formulated as $q\left (\mathbf{g}_t \mid \mathbf{g}_{t-1}\right ) = \mathcal{N}\left ( \sqrt{1-\beta_t}\mathbf{g}_{t-1}, \beta_t\mathbf{I}\right )$. The noisy sample at any arbitrary time step $t$ can be obtained in a closed form of:
\begin{equation}
    \mathbf{g}_t = \sqrt{\bar{\alpha}_t}\mathbf{g}_0 + \sqrt{1-\bar{\alpha}_t}\bm{\epsilon},
\end{equation}
where $\bar{\alpha}_t =\prod_{i=1}^{t}\alpha_t$ with $\alpha_t = 1-\beta_t$ and $\bm{\epsilon} \sim \mathcal{N}\left (\mathbf{0}, \mathbf{I} \right )$.

\textbf{Reverse Process.}
The reverse process allows us to generate a pose from the Gaussian noise by gradually denoising through $T$ steps via the reverse probability $q\left (\mathbf{g}_{t-1} \mid \mathbf{g}_t, \mathbf{c}, \mathbf{t} \right )$. In this probability, $\mathbf{c}$ is the point cloud feature produced by the PointNet++ encoder and $\mathbf{t}$ is the text embedding of the affordance of interest. $\mathbf{c}$ and $\mathbf{t}$ represent two guidances that our model need to condition on, i.e., the point cloud object and the affordance text. As $q\left (\mathbf{g}_{t-1} \mid \mathbf{g}_t, \mathbf{c}, \mathbf{t} \right )$ is intractable~\cite{ho2020denoising}, we approximate it with a neural network. More particularly, we approximate the noise $\bm{\epsilon}$ at every timestep $t$ by a denoising network $\bm{\epsilon}_{\bm{\theta}}\left (\mathbf{g}_t, \mathbf{c},\mathbf{t}, t \right )$. $\bm{\epsilon}$ is updated to minimize the difference between the real and approximated noises. The loss function for pose generation is therefore computed as:
\begin{equation}
    \mathcal{L}_{\text{pose}} = \mathbb{E}_{\bm{\epsilon}, \mathbf{g}_0, \mathbf{c}, \mathbf{t}, t}\left [ \left \| \bm{\epsilon} - \bm{\epsilon}_{\bm{\theta}} \left ( \mathbf{g}_t, \mathbf{c},\mathbf{t}, t \right ) \right \|^2\right].
\end{equation}
Following other works~\cite{ho2022classifier}, to balance between the quality and the diversity of the generated poses, we randomly drop the conditions $\mathbf{c}$ and $\mathbf{t}$ to train unconditionally with a probability $p_{\text{uncond}}$. Our design allows conditional training and unconditional training to use a single network.

Subsequently, we detail the design of our denoising network $\bm{\epsilon}_{\bm{\theta}}$. Kindly refer to Figure~\ref{fig:architecture} for an illustrative demonstration. In particular, following other works of diffusion models~\cite{ho2020denoising}, we employ a downscale-upscale U-Net architecture~\cite{ronneberger2015u} for the network. The noisy pose $\mathbf{g}_t$ at timestep $t$ is first plugged into three consecutive downscaling MLPs, and then, three other consecutive MLPs are used in the upscaling phase. To form the input to each upscaling MLP, we combine the output of the preceding one, the feature from skip connection, the time embedding $\mathbf{\tau}$ computed from the timestep $t$, and the unified context $\mathbf{u}$ combining the point cloud feature $\mathbf{c}$ and the text feature $\mathbf{t}$. The unified context $\mathbf{u}$ is obtained via a ContextNet module. In this ContextNet, we first compute the point cloud influence mask $\mathbf{m}_{\mathbf{c}}$ and text influence mask $\mathbf{m}_{\mathbf{t}}$ using two MLPs and a softmax layer. The influence masks are at the same size as the two features. The unified context $\mathbf{u}$ is then computed as:
\begin{equation}
    \mathbf{u} = \mathbf{c} \odot \mathbf{m}_{\mathbf{c}} + \mathbf{t} \odot \mathbf{m}_{\mathbf{t}},
\end{equation}
where $\odot$ represents the element-wise multiplication.

\textbf{Pose Sampling.}
When finishing the model training, we can sample poses from Gaussian noise by applying the reverse process from timestep $T$ to $0$ using the formulation:
\begin{equation}
    \mathbf{g}_{t-1} = \frac{1}{\sqrt{\alpha_t}}\left ( \mathbf{g}_t - \frac{1-\alpha_t}{\sqrt{1-\bar{\alpha}_t}}\bar{\bm{\epsilon}}_{\bm{\theta}}\left ( \mathbf{g}_t, \mathbf{c}, \mathbf{t}, t \right ) \right ) + \sqrt{\beta_t}\mathbf{z},
\end{equation}
where $\mathbf{z} \sim \mathcal{N}\left (\mathbf{0}, \mathbf{I} \right )$ if $t>1$, else $\mathbf{z} = \mathbf{0}$, and $\bar{\bm{\epsilon}}_{\bm{\theta}}\left ( \mathbf{g}_t, \mathbf{c}, \mathbf{t}, t \right )$ is calculated as:
\begin{equation}
    \bar{\bm{\epsilon}}_{\bm{\theta}}\left ( \mathbf{g}_t, \mathbf{c}, \mathbf{t}, t \right ) = (w+1){\bm{\epsilon}}_{\bm{\theta}}\left ( \mathbf{g}_t, \mathbf{c}, \mathbf{t}, t \right ) - w\bm{\epsilon}_{\bm{\theta}}\left ( \mathbf{g}_t, t \right ),
\end{equation}
where $w$ is a guidance scale hyperparameter and $\bm{\epsilon}_{\bm{\theta}}\left ( \mathbf{g}_t, t \right )$ is the predicted noise when the conditions are discarded.

\subsection{Training and Inference}
We define the overall loss function as  $\mathcal{L} = \mathcal{L}_{\text{aff}} + \mathcal{L}_{\text{pose}}$. The number of points in each point cloud is fixed to $n=2,048$. We utilize the state-of-the-art CLIP text encoder~\cite{radford2021learning} and freeze it during training. For the diffusion model, we set $T=1,000$, and set the forward diffusion variances to constants
increasing linearly from $\beta_1 = 10^{-4}$ to $\beta_T=0.02$. The unconditional training probability is set to $p_{\text{uncond}} = 0.05$. The whole network is trained end-to-end over 200 epochs on a 24GB-RAM NVIDIA GeForce RTX 3090 Ti with a batch size of 32. The Adam optimizer~\cite{kingma2014adam} with the learning rate $10^{-3}$ and the weight decay $10^{-4}$ is used. When sampling poses, we set the guidance scale to $w=0.2$. Our framework takes \SI{180}{\milli\second} to detect affordances and generate $2,000$ corresponding 6-DoF poses for one point cloud.

\section{Experiments} \label{Sec:exp}
In this section, we conduct several experiments to demonstrate the effectiveness of our proposed 3DAPNet trained on our 3DAP dataset. We start by comparing our method with other baselines
. Second, we present 3DAPNet's notable qualitative results. Third, we provide different ablation studies for a more in-depth investigation of our method. Finally, we validate our framework in real robotic experiments.      
\subsection{Quantitative Comparisons}
\textbf{Baselines.}
We compare our 3DAPNet with the following methods: 6D-TGD\cite{chen2022learning}, OpenAD~\cite{nguyen2023open}, 6D-GraspNet~\cite{mousavian20196}. Note that, OpenAD does not support pose estimation, while 6D-GraspNet does not support affordance detection. We tailor 6D-GraspNet~\cite{mousavian20196} to open-vocabulary setting by incorporating the affordance text branch to the network input. 
All methods are trained on our dataset with the splitting ratio for training, evaluation, and testing of 7:1:2.

\textbf{Metrics.} For affordance detection, following~\cite{nguyen2023open}, we evaluate the methods using three metrics, i.e., mIoU (mean IoU over all affordance classes), Acc (overall accuracy of all points), and mAcc (mean accuracy over all affordances). For pose generation, we use two metrics as in recent works: the mean evaluated similarity metric (mESM)~\cite{chen2022learning} and the mean coverage rate (mCR)~\cite{mousavian20196}. 
To validate the pose estimation results, we generate 200 poses for each pair of object-affordance during the testing.

\begin{figure}[t]
	\centering
	\includegraphics[width=0.85\linewidth]{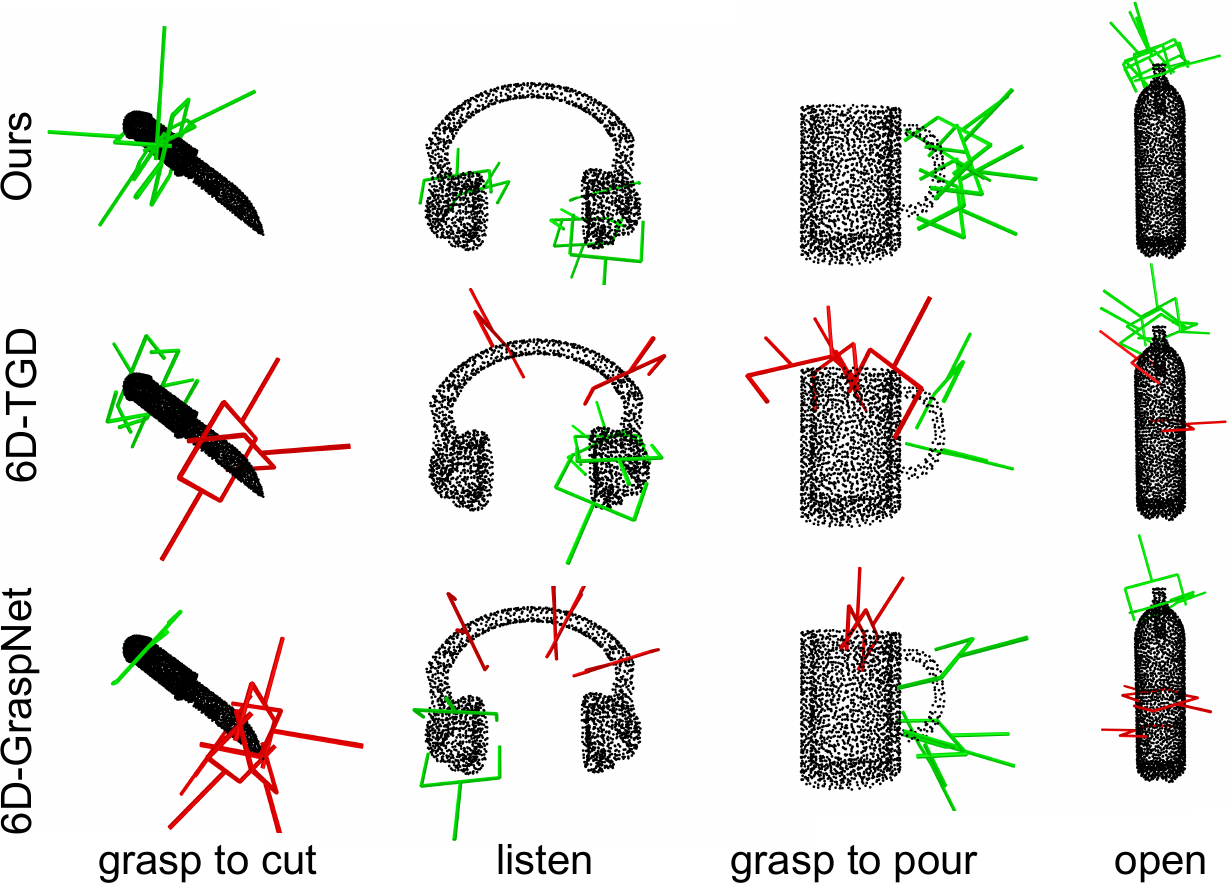}
 \vspace{0.5ex}
	\caption{Qualitative comparison. \textcolor[HTML]{00FF00}{Green} color denotes poses that are related to the input text, while~\textcolor[HTML]{FF0000}{red} indicates poses not related to the input text.}
	\label{fig:qualitative_compare}
\end{figure}

\begin{table}[t]
\caption{Baseline comparisons}
\label{tab: baseline_compare}
\vskip 0.15in
\begin{tabular}{lccccc}
\toprule
\multirow{2}{*}{Method} & \multicolumn{3}{c}{Affordance Detection}                   & \multicolumn{2}{c}{Pose Estimation} \\ \cmidrule(lr){2-4} \cmidrule(lr){5-6} 
                        & mIoU$\uparrow$ & Acc$\uparrow$ & mAcc$\uparrow$ & mESM$\downarrow$   & mCR$\uparrow$    \\ \midrule
OpenAD~\cite{nguyen2023open}                  & 55.20     & 58.89    & 58.22     & --       & --      \\
6D-GraspNet~\cite{mousavian20196}                & --  & --   & --    &  0.373       & 22.81       \\
6D-TGD~\cite{chen2022learning}                  & 55.38     & 60.14    & 57.99     & 0.219        & 30.10       \\
3DAPNet (ours)      & \bf{56.18}     & \bf{61.77}     & \bf{59.26}      & \bf{0.120}        & \bf{44.63}       \\ \bottomrule
\end{tabular}
\end{table}

\textbf{Result.}
Table~\ref{tab: baseline_compare} illustrates the performance of our 3DAPNet across both tasks, consistently achieving the highest scores across all five metrics when compared to alternative approaches. Specifically, in the affordance detection task, 3DAPNet exhibits improvements of 0.8\% on mIoU, 1.63\% on Acc, and 1.04\% on mAcc relative to its closest competitors. Regarding pose estimation, 3DAPNet is nearly twice as good as the runner up 6D-TGD on mESM metric (0.120 compared to 0.219). Furthermore, 3DAPNet significantly outperforms competitors in the mCR metric, with a score of 44.63\% compared to the second-best 6D-TGD's 30.10\%.

\subsection{Qualitative Results}

\textbf{Qualitative Comparison.}
We present qualitative results to compare our 3DAPNet with other methods in pose generation capability. Particularly, we select poses generated by our method, 6D-GraspNet~\cite{mousavian20196}, and 6D-TGD~\cite{chen2022learning}. The example poses are shown in Figure~\ref{fig:qualitative_compare}. We observe that our method produces more poses that directly support the affordance tasks, while in contrast, the other two baselines generate a large number of poses that do not facilitate them. This result further highlights the enhanced effectiveness of our approach.

\begin{figure}[t]
	\centering
	\includegraphics[width=1.0\linewidth]{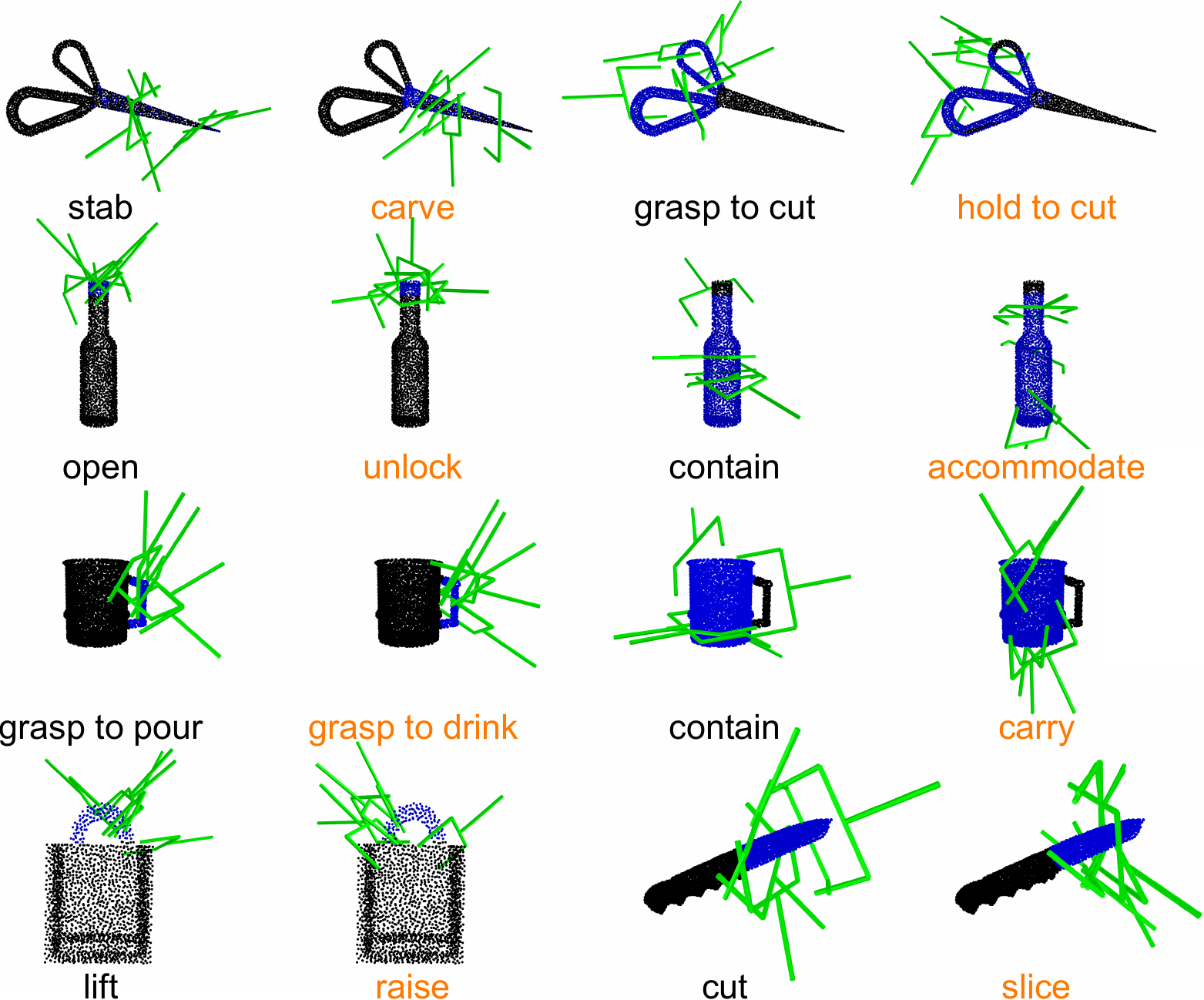}
 \vspace{0.5ex}
	\caption{Qualitative results of 3DAPNet's generalization to unseen affordances. The unseen affordances are shown in \textcolor[HTML]{FF7800}{orange}.}
	\label{fig:unseen_affordance}
\end{figure}

\textbf{Generalization to Unseen Affordances.}
We present several examples demonstrating 3DAPNet's ability to generalize to unseen affordances in Figure~\ref{fig:unseen_affordance}. For affordances in the training set, 3DAPNet yields high-quality results both in affordance detection and pose generation. With the reference of seen affordances, when evaluating on unseen affordances, our method still succeeds in detecting the associated regions and generating the corresponding appropriate poses, though those affordances do not appear in the training set.

\begin{figure}[h]
	\centering
	\includegraphics[width=0.8\linewidth]{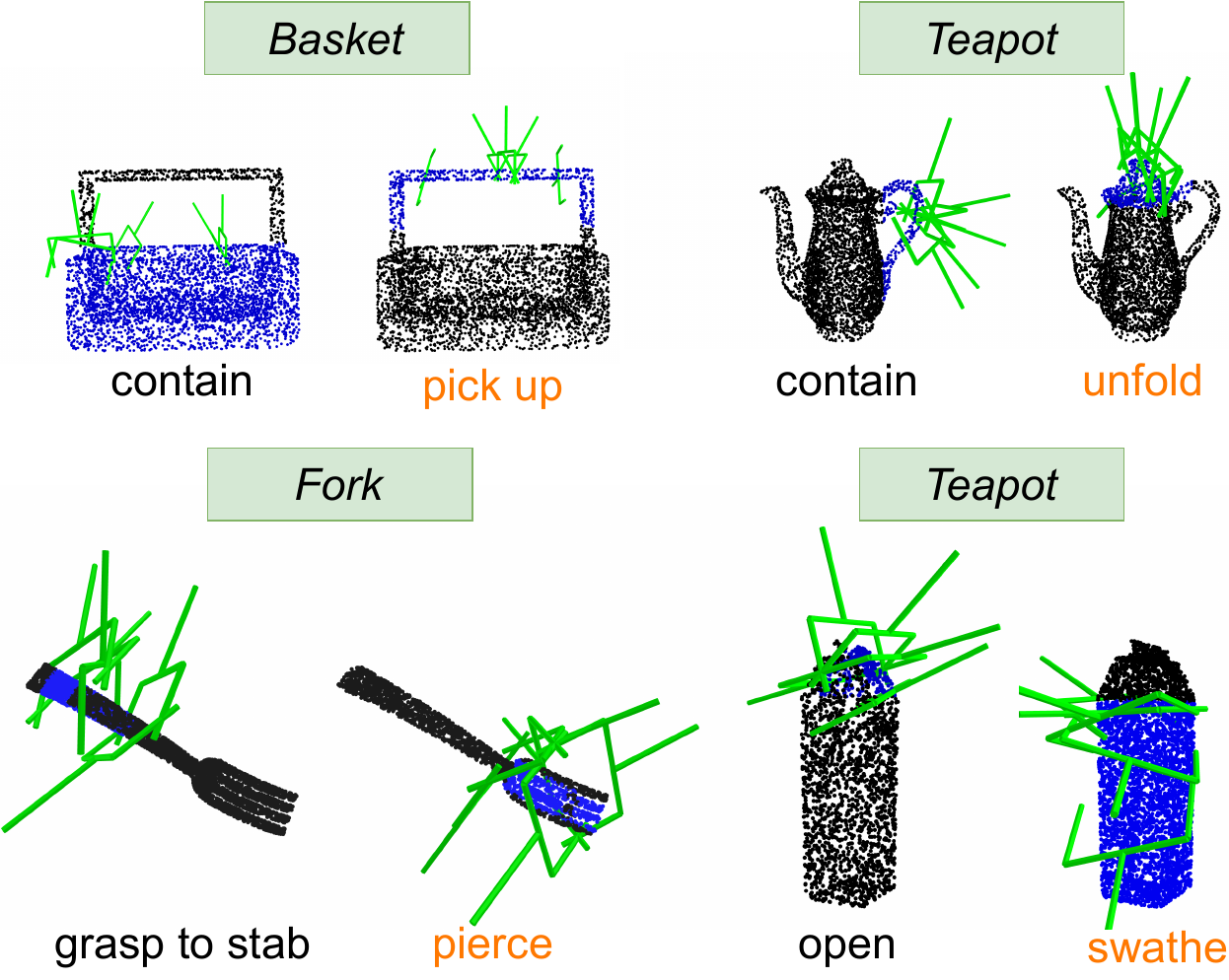}
 \vspace{0.5ex}
	\caption{Qualitative results of 3DAPNet's generalization to unseen object categories. The unseen affordances are shown in~\textcolor[HTML]{FF7800}{orange}.}
	\label{fig:unseen_object}
\end{figure}

\textbf{Generalization to Unseen Objects.}
We extend our assessment to the broader context of unseen object categories. Concretely, we curate new objects from the ShapeNetCore dataset~\cite{chang2015shapenet} and feed both seen and unseen affordances to the model.
The reasonable outcomes shown in Figure~\ref{fig:unseen_object} reaffirm the generalization of our 3DAPNet.

\subsection{Ablation Study}

\textbf{Single branch vs. jointly learned network.} We investigate the performance of each branch in our 3DAPNet on its corresponding task while excluding the other. The results are detailed in Table~\ref{tab: single-branch}. In the case of pose generation only, we retain the usage of the point cloud and text embeddings by keeping the PointNet++ encoder and the CLIP text encoder, while remove the PointNet++ decoder and the correlation head. The results indicate that the combination of the two branches yields the highest performance on both tasks when compared to each branch operating individually. This further validates the efficacy of our design, where the learning processes of the two tasks mutually benefit each other.

\begin{table}[t]
\caption{Single-Branch vs. Jointly Learned Network}
\label{tab: single-branch}
\vskip 0.15in
\begin{tabular}{C{2.0cm}ccccc}
\toprule
\multirow{2}{*}{Method} & \multicolumn{3}{c}{Affordance Detection}                   & \multicolumn{2}{c}{Pose Estimation} \\ \cmidrule(lr){2-4} \cmidrule(lr){5-6} 
                        & mIoU$\uparrow$ & Acc$\uparrow$ & mAcc$\uparrow$ & mESM$\downarrow$   & mCR$\uparrow$    \\ \midrule
Affordance Only                  & 55.65 & 60.73 & 58.80 & --      & --     \\
Pose Only                & -- & --  & -- & 0.147 & 41.29 \\
Both                & \bf{56.18}     & \bf{61.77}     & \bf{59.26}      & \bf{0.120}        & \bf{44.63}       \\
\bottomrule
\end{tabular}
\end{table}

\textbf{Effectiveness of ContextNet.}
\begin{table}[t]
\caption{The Effectiveness of ContextNet}
\label{tab: contextnet}
\vskip 0.15in
\begin{tabular}{C{2.0cm}ccccc}
\toprule
\multirow{2}{*}{ContextNet} & \multicolumn{3}{c}{Affordance Detection}                   & \multicolumn{2}{c}{Pose Estimation} \\ \cmidrule(lr){2-4} \cmidrule(lr){5-6} 
                        & mIoU$\uparrow$ & Acc$\uparrow$ & mAcc$\uparrow$ & mESM$\downarrow$   & mCR$\uparrow$    \\ \midrule
\XSolidBrush                  & 53.97    & 60.20   & 58.49     & 0.433      & 12.61     \\
\Checkmark                & \bf{56.18}     & \bf{61.77}     & \bf{59.26}      & \bf{0.120}        & \bf{44.63}       \\
\bottomrule
\end{tabular}
\end{table}
We further validate the effectiveness of our framework design by performing an ablation experiment on the ContextNet. Specifically, we report the performances of our framework with and without the ContextNet in Table~\ref{tab: contextnet}. In the case of ContextNet is removed, we combine the point cloud and affordance text conditions naively by adding them. The empirical result shows that our framework with ContextNet completely dominates the other one, with a very large gap on the task of pose generation, which is the main target of our design.

\begin{table}[h]
\caption{Text Encoder Analysis}
\label{tab: text_encoder}
\vskip 0.15in
\begin{tabular}{C{2.0cm}ccccc}
\toprule
\multirow{2}{*}{Text Encoder} & \multicolumn{3}{c}{Affordance Detection}                   & \multicolumn{2}{c}{Pose Estimation} \\ \cmidrule(lr){2-4} \cmidrule(lr){5-6} 
                        & mIoU$\uparrow$ & Acc$\uparrow$ & mAcc$\uparrow$ & mESM$\downarrow$   & mCR$\uparrow$    \\ \midrule
BERT~\cite{devlin2018bert}                  &  52.89   & 59.77   &  59.25    & 0.182      & 30.46     \\
RoBERTa~\cite{liu2019roberta}                &  53.92  & 58.13   &  57.01   &  0.277    & 30.11       \\
CLIP~\cite{radford2021learning}                  & \bf{56.18}     & \bf{61.77}     & \bf{59.26}      & \bf{0.120}        & \bf{44.63}       \\ \bottomrule
\end{tabular}
\end{table}

\textbf{Text Encoder.}
As the text encoder is critical in our framework, we conduct extensive study to investigate the performances of different text encoders. In particular, we use three state-of-the-art text encoder, which are BERT~\cite{devlin2018bert}, RoBERTa~\cite{liu2019roberta}, and CLIP~\cite{radford2021learning}. The result is shown in Table~\ref{tab: text_encoder}. We observe that the CLIP encoder significantly outperforms its counterparts on both tasks, especially on pose generation. This result demonstrates the superiority of CLIP in language-vision understanding.

\begin{figure}[t]
\centering
\subfigure{
\label{fig:rotbot1}
\def\svgwidth{0.9\columnwidth}
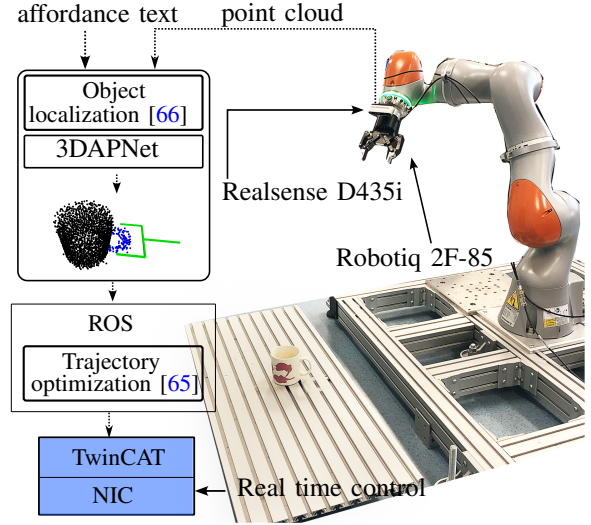
}
\vspace{1ex}
\caption{The overview of the robot experiment setup. More qualitative results can be found in our Demonstration Video.}
\label{fig: robot demonstration}
\end{figure}

\begin{figure}[h]
	\centering
	\includegraphics[width=0.9\linewidth]{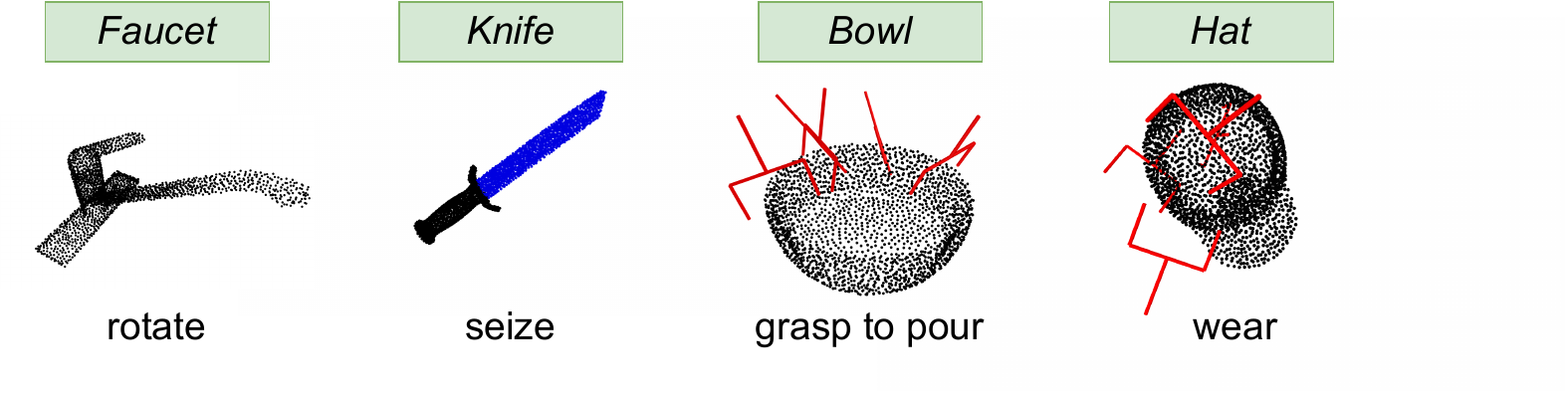}
 \vspace{1ex}
	\caption{Some wrong cases of our method.}
	\label{fig:wrong_cases}
\end{figure}

\subsection{Robotic Demonstration}

The experiment setup, shown in Figure~\ref{fig: robot demonstration}, comprises three main modules, i.e., the inference module, the ROS module, and the real-time controller module.
In the inference module, after receiving point cloud data of the environment from the RealSense D435i camera, we utilize the state-of-the-art object localization method~\cite{zhou2018voxelnet} to identify the object, then perform point sampling to get $2,048$ points.  
We then feed this point cloud with a text affordance command into our proposed 3DAPNet. 
Then, the generated affordance pose from 3DAPNet is sent to ROS for planning and trajectory generation. Analytical inverse kinematics \cite{vu2023machine} and trajectory optimization \cite{beck2022singularity} are employed to compute optimal trajectories of the robot to reach the computed pose provided by our network. 
Note that, using our 3DAPNet, we can have a general input command and are not restricted to a predefined affordance label set. 
Several demonstrations can be found in our Demonstration Video.

\subsection{Discussion}

Despite promising results, it is important to acknowledge that our method has not fulfilled the perfect ability in universal affordance detection and pose estimation. There are cases where our method shows its limitations, which are presented in Figure~\ref{fig:wrong_cases}. Particularly, on the left are two cases of fail and false-positive detection of unseen affordances. In two cases on the right, we show examples where our method generates poses that do not facilitate the corresponding affordances. Furthermore, our method can only detect affordance from single objects due to the dataset limitation. This leads to the fact that it is not straightforward to perform evaluation on real robots with other methods. Therefore, having a large-scale dataset with cluttered point cloud scenes would enable  more qualitative comparisons and applications.

\section{Conclusions}\label{Sec:con}
We have tackled the task of open-vocabulary affordance detection and pose estimation in 3D point clouds. In particular, we have presented the 3DAP dataset for affordance-pose joint learning and proposed the 3DAPNet method that can simultaneously detect open-vocabulary affordances and generate affordance-specific 6-DoF poses. Experimental results show that our approach outperforms other methods by a large margin on both tasks. We extensively demonstrated the effectiveness of 3DAPNet in real-world robotic manipulation applications. We hope that the prospective results of our 3DAPNet could encourage more future researchers to further investigate this important yet challenging problem. Our code and trained model will be made publicly available. 

\bibliographystyle{class/IEEEtran}
\bibliography{class/IEEEabrv,class/reference}
   
\end{document}